\documentclass[11pt]{article}

\usepackage[preprint]{acl}

\usepackage{times}
\usepackage{latexsym}
\usepackage{booktabs}

\usepackage[T1]{fontenc}

\usepackage[utf8]{inputenc}

\usepackage{microtype}

\usepackage{inconsolata}

\usepackage{graphicx}

\title{3D-TAFS: A Training-free Framework for 3D Affordance Segmentation
}

\author{Meng Chu\\
  HKUST\\\And
  Xuan Zhang\\
  NUS\\\And
  Zhedong Zheng\\
  University of Macau\\\And
  Tat-Seng Chua \\
  NUS\\
  }

\begin{document}
\maketitle
\begin{abstract}
Translating high-level linguistic instructions into precise robotic actions in the physical world remains challenging, particularly when considering the feasibility of interacting with 3D objects. In this paper, we introduce 3D-TAFS, a novel training-free multimodal framework for 3D affordance segmentation. To facilitate a comprehensive evaluation of such frameworks, we present IndoorAfford-Bench, a large-scale benchmark containing 9,248 images spanning 20 diverse indoor scenes across 6 areas, supporting standardized interaction queries. In particular, our framework integrates a large multimodal model with a specialized 3D vision network, enabling a seamless fusion of 2D and 3D visual understanding with language comprehension. Extensive experiments on IndoorAfford-Bench validate the proposed 3D-TAFS's capability in handling interactive 3D affordance segmentation tasks across diverse settings, showcasing competitive performance across various metrics. Our results highlight 3D-TAFS's potential for enhancing human-robot interaction based on affordance understanding in complex indoor environments, advancing the development of more intuitive and efficient robotic frameworks for real-world applications. 

\end{abstract}

\section{Introduction}

In the rapidly evolving field of robotics and computer vision, the ability to understand and interact with complex 3D environments remains a frontier ripe for exploration. Recent years have witnessed unprecedented advancements in artificial intelligence, particularly with the emergence of large language models (LLMs) and vision-language models  \cite{kim2024openvla, wei2024incorporating, hong2023fluxformer}. These breakthroughs have revolutionized numerous aspects of AI, from natural language processing to image recognition. However, a significant challenge persists: bridging the gap between high-level linguistic instructions and precise 3D robotic actions in real-world scenarios \cite{chen2023polarnet, tziafas2023language, ahn2022can}.

The integration of language understanding with spatial reasoning and manipulation skills is crucial for the next generation of intelligent frameworks \cite{ha2023scaling, zhang2023noir}. While LLMs excel at processing and generating human-like text, and vision models can interpret complex visual scenes, translating this understanding into actionable 3D interactions remains an open problem. This challenge is particularly evident in embodied AI applications, where agents must navigate, manipulate, and interact with their physical surroundings based on natural language instructions \cite{xia2024kinematic, zhou2024isr}.

\begin{figure}[t]
  \centering
      \includegraphics[width=0.45\textwidth]{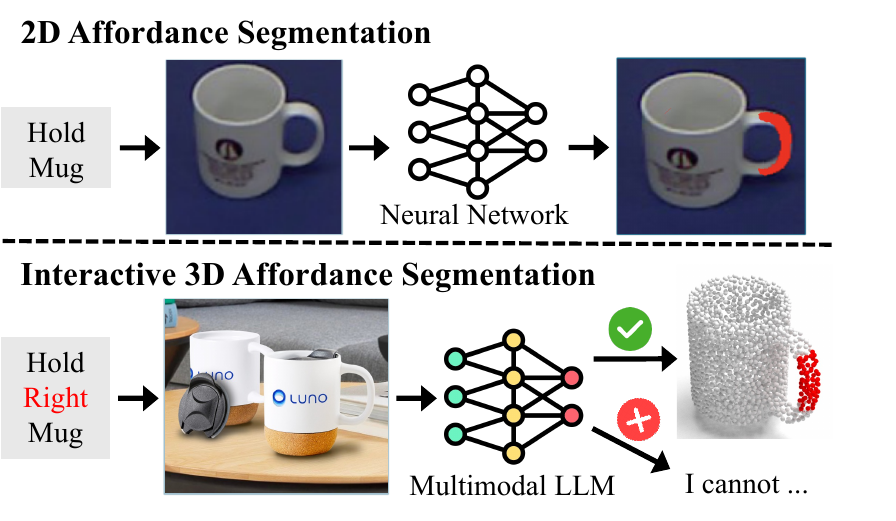}
  \caption{\textbf{Comparison of 2D affordance segmentation and interactive 3D affordance segmentation.} While 2D segmentation offers simplicity for static image analysis, interactive 3D segmentation introduces interactivity, multimodal processing, and richer spatial understanding.
  }
  \label{fig:head}
\end{figure}

Traditional approaches in robotics and computer vision have often addressed 2D and 3D domains separately, lacking the holistic perspective necessary for effective embodied interaction \cite{xing2024masked, vidanapathirana2023spectral}. Two-dimensional visual understanding, while advanced, falls short of capturing the full complexity of real-world environments. Conversely, pure 3D approaches often struggle with semantic interpretation and language grounding \cite{liu2024lps, wang2024joint}. This dichotomy has limited the development of truly versatile and intuitive robotic frameworks capable of understanding and acting upon nuanced human instructions in diverse settings \cite{ausserlechner2024zs6d,9758071}.

Recent research has begun to explore the potential of LLMs in embodied navigation and planning tasks \cite{pmlr-v205-shah23b, shah2023navigation}. These studies have shown promising results in high-level decision-making and route planning. However, they frequently encounter limitations in fine-grained manipulation tasks that demand precise spatial understanding and object interaction \cite{long2024novel}. The ability to grasp the affordances of objects—their potential uses and interactions—in a 3D context while aligning with natural language instructions remains a significant hurdle \cite{lu2023ovir}.

\begin{figure*}[t]
  \centering
\includegraphics[width=\textwidth]{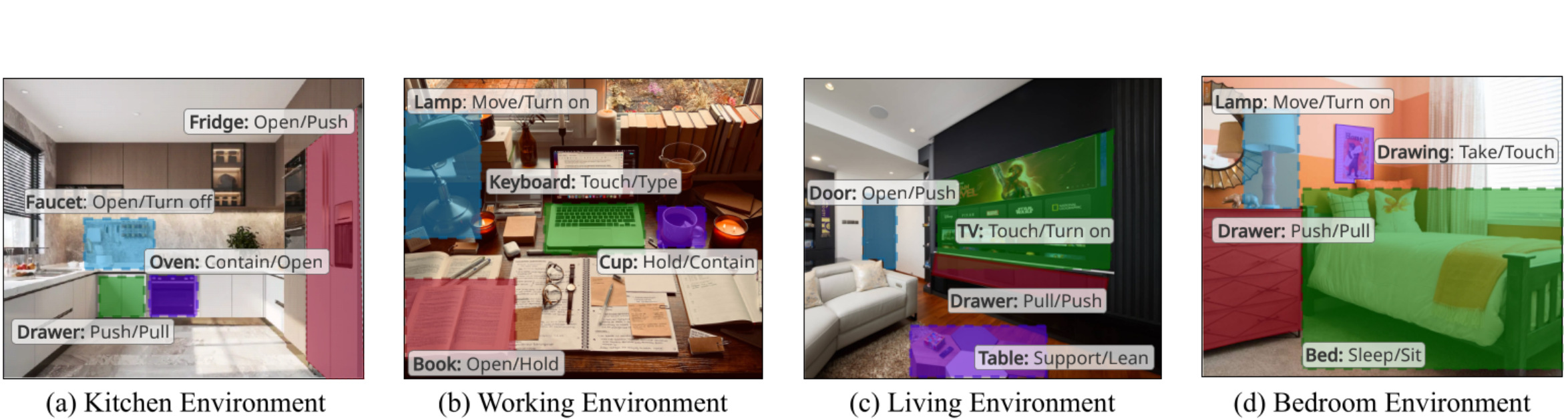}
  \caption{\textbf{Demonstration of possible affordance in different environments.} This image provides a comprehensive overview of human-object interactions across four common domestic environments: kitchen, working space, living room, and bedroom. By mapping out specific objects in each space and their associated actions, it offers valuable insights into how people engage with their surroundings daily.  }
  \label{fig:environment}
\end{figure*}
To address these challenges, we propose 3D-TAFS, a novel training-free multimodal framework for 3D affordance segmentation. As shown in Figure \ref{fig:head}, our framework is designed to bridge the gap between semantic and spatial comprehension, enabling more intuitive and effective human-robot interaction. The core motivation behind our framework stems from the need to equip embodied agents with the capability to seamlessly integrate 2D and 3D visual understanding with language comprehension \cite{zitkovich2023rt}.

3D-TAFS leverages the strengths of large multimodal models, combining them with specialized networks to process and reason about visual and linguistic inputs in tandem \cite{zitkovich2023rt, saxena2023multi, GeoText1652}. This integration allows our framework to perform sophisticated vision-language reasoning, translating high-level instructions into precise 3D affordance segmentation without additional training. By doing so, our framework opens new possibilities for robots to understand and interact with their environment in ways that more closely align with human intentions and expectations \cite{kim2024openvla, chen2023polarnet}.

3D-TAFS tackles these limitations by uniquely combining two effective components: a large multimodal model for vision-language understanding \cite{zhang2024next} and a specialized network for language-guided 3D affordance segmentation \cite{Li_2024_CVPR}.
This integration enables our framework to process multimodal inputs, perform vision-language understanding, localize objects, retrieve and register 3D point clouds, and execute language-guided 3D affordance segmentation without additional training.
The key contributions of this work are as follows:
\begin{itemize}

    \item We present a training-free multimodal framework linking high-level instructions and precise robotic actions in 3D environments. 
    Specifically, our framework
    integrates 2D and 3D visual understanding with language comprehension for embodied agents. 
    
\item We introduce IndoorAfford-Bench, a large-scale indoor scene-object-affordance relationship benchmark containing 9,248 images across 20 diverse indoor scenes in 6 areas. The dataset provides rich annotations for 22 object categories and 18 affordance types, enabling comprehensive evaluation of 3D affordance understanding with 180 standardized interaction queries.
    
    \item To evaluate interactive language-guided affordance segmentation in everyday environments, we provide a new dataset, containing comprehensive testing and development of multimodal frameworks for complex spatial understanding tasks. 
    Our proposed method has achieved state-of-the-art performance in 3D affordance analysis and segmentation across diverse indoor environments.
\end{itemize}

\section{Related Work} \label{sec:rel-works}

\noindent\textbf{Large Models for Visual Understanding.} 
Large models have significantly trumped visual understanding tasks with the supervision of language \cite{huang2023language,li2023blip}. In 2D visual grounding, GPT4ROI \cite{zhang2023gpt4roi} encodes region features interleaved with language embeddings for fine-grained multimodal reasoning. Shikra \cite{chen2023shikra} further improves visual grounding in the unified natural language form. 3D environments bring out greater complexity but provide more precise details than 2D images. For 3D understanding, Chen et al. established ScanRefer \cite{chen2020scanrefer} to learn the correlated representation between 3D object proposals and encoded description embeddings. Building on this, ScanQA \cite{azuma2022scanqa} is formulated for 3D question answering. However, these methods only focus on either 2D or 3D domains separately, lacking the holistic perspective for embodied agents. 3D-TAFS bridges this gap by seamlessly combining 2D and 3D visual understanding with language comprehension.

\noindent\textbf{Embodied Agents for Robotic Tasks.}
Embodied agents in robotics aim to unify visual perception and physical action in real-world environments. To enable and encourage the application of situated multimodal learning, vision-and-language navigation \cite{anderson2018vision} is first presented for embodied learning. Furthermore, Hong et al.  \cite{hong2020recurrent} equips the BERT model recurrent functions to capture the cross-model time-aware information for agents. As the planning capability of Large Language Models (LLMs) has revolutionized the vision-language problem \cite{brown2020language, jiang2023mistral}, some researchers attempt to apply LLMs as an auxiliary module for embodied navigation. Huang et al. \cite{huang2022inner} extends the powerful reasoning ability of LLMs grounding on embodied context and language feedback. In contrast, Singh et al. \cite{singh2023progprompt} structures program-like prompts to enable the universal plan generation across diverse situated tasks. Chen et al. \cite{chen2024mapgpt} build an online language-formed map to extend the agent action space from local to global. These works highlight the potential of language models in embodied robot planning, yet they often struggle with fine-grained manipulation that requires precise spatial understanding. Our paper addresses this limitation by integrating LLMs with 3D point cloud processing for highly accurate object interaction and manipulation.

\begin{figure*}[t]
  \centering
  \includegraphics[width=0.9\textwidth]{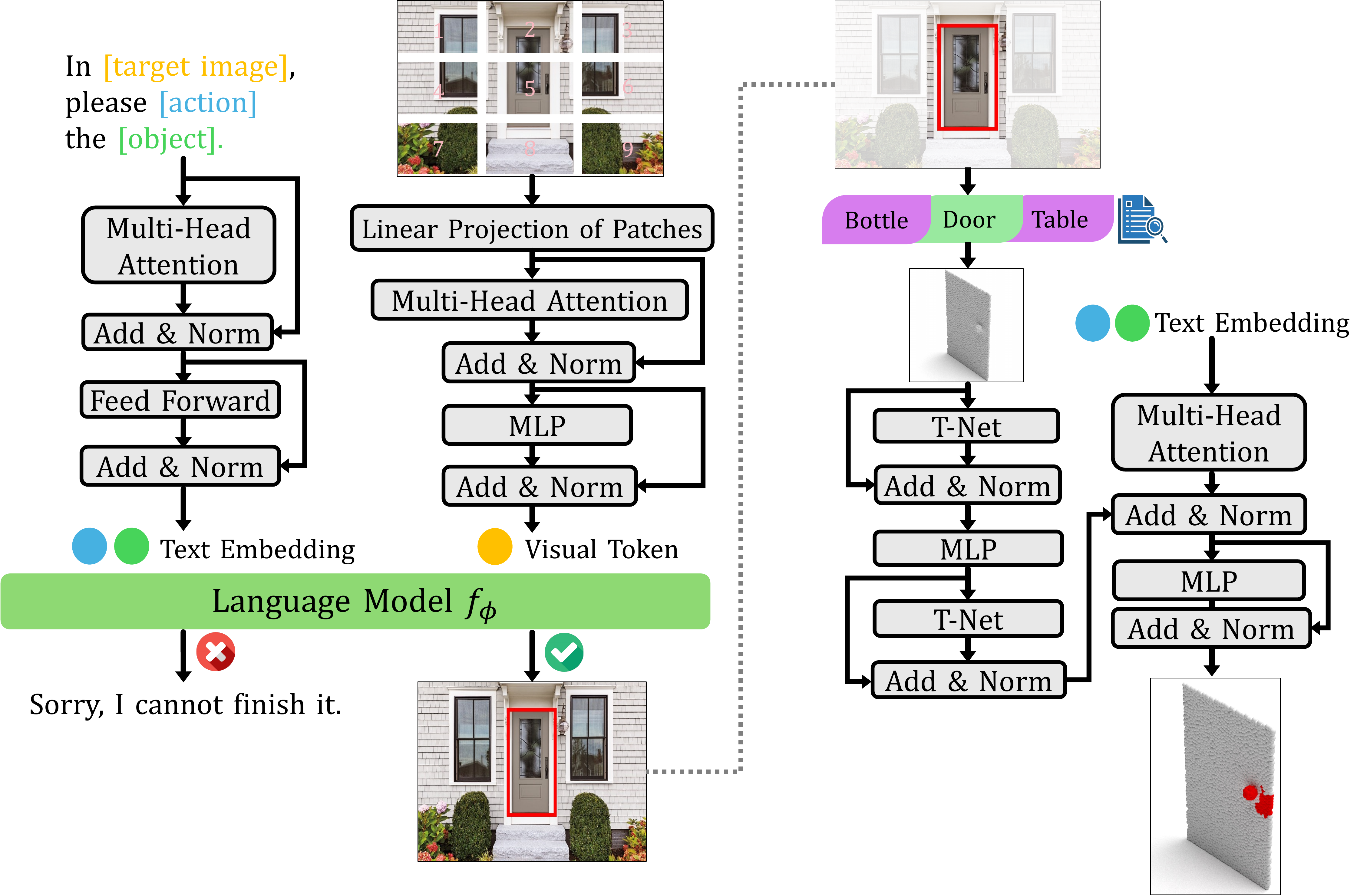}
  \caption{\textbf{Sturcture and working flow of 3D-TAFS.} Our framework integrates vision-language processing with 3D affordance segmentation for robotic action guidance. It depicts two parallel input streams: visual input undergoing linear projection and multi-head attention and textual input processing through multi-head attention and feed-forward networks. These streams converge in a language model, enabling cross-modal understanding. Then, it decides to do object label identification to find the standard 3D point cloud. Finally, the framework starts to do the 3D affordance segmentation. This architecture demonstrates the seamless integration of computer vision, natural language processing, and robotics to create a sophisticated framework capable of understanding and interacting with its environment in a human-like manner.}
  \label{fig:main}
\end{figure*}

\noindent\textbf{Affordance Learning in Robotics.}
Affordance learning is crucial for robotic manipulation tasks. Traditional approaches like 3D AffordanceNet \cite{do2018affordancenet} focused on learning affordances by detecting objects in the end-to-end architecture. Yang et al. \cite{yang2023grounding} extend this by proposing a setting for learning 3D affordance parts guided by image demonstrations but discarding the semantic information. Recently, Li et al. \cite{Li_2024_CVPR} introduced PointRefer, a novel task for language-guided affordance segmentation on 3D objects. While these works have made significant strides in affordance detection, they often lack the flexibility to integrate with diverse, context-rich instructions under LLM generation.

Our approach differs by directly learning from linguistic context, aligning more closely with the semantic richness of LLMs and their potential downstream applications in robotics. As shown in Figure~\ref{fig:environment}, however, humans typically perceive and communicate about their environment in 2D, while robots need to perform tasks in 3D spaces with precise actions. 3D-TAFS bridges this gap by interpreting 2D visual information from humans and translating it into 3D actions for robots. This capability is crucial as robots become more integrated into our daily lives, from homes to factories.

\section{Methodology}\label{sec:methodology}
We present 3D-TAFS, a novel training-free multimodal framework for advanced object understanding and interaction. Our approach integrates a large multimodal model for vision-language understanding and a specialized network for language-guided 3D affordance segmentation.

\subsection{Framework Overview}
Figure~\ref{fig:main} illustrates a comprehensive framework for language-guided robotic interactions, comprising several key stages. The process begins with multimodal input processing of visual and textual information, followed by vision-language understanding and object localization to interpret the input and identify relevant objects. Next, 3D point cloud retrieval and registration align 2D visual data with 3D spatial information. The fourth stage involves language-guided 3D affordance segmentation, determining how objects can be interacted with based on given instructions. Finally, 2D and 3D information are integrated for the final output, bridging the gap between high-level commands and precise robotic actions. This approach allows the framework to understand complex instructions and translate them into actionable insights for robotic frameworks, enabling accurate, language-guided interactions in 3D environments.

\begin{figure*}[t]
\centering
\begin{minipage}{0.1\textwidth}  
{\small
\begin{tabular}{lr}
    \toprule
    \multicolumn{2}{l}{\textbf{Basic Statistics}} \\
    \midrule
    Total number of scenes & 20 \\
    ~- Total number of objects & 22 \\
    ~- Total number of affordances & 18 \\
    ~- Total number of affordance queries & 180 \\
    ~- Total number of images & 9,248 \\
    ~- Number of data sources & 6 \\
    \midrule
    \multicolumn{2}{l}{\textbf{Average Statistics}} \\
    \midrule
    Average objects per scene & 19.35 \\
    Average affordances per scene & 16.75 \\
    Average affordance queries per scene & 167.50 \\
    Average images per scene & 462.40 \\
    Average objects per image & 5.46 \\
    Average affordances per image & 16.12 \\
    Average affordance queries per image & 161.20 \\
    \midrule
    \multicolumn{2}{l}{\textbf{Object and Affordance Distribution}} \\
    \midrule
    Maximum objects per image & 35 \\
    Minimum objects per image & 1 \\
    Maximum affordances per image & 175 \\
    Minimum affordances per image & 1 \\
    \midrule
    \multicolumn{2}{l}{\textbf{Data Source Distribution}} \\
    \midrule
    Houzz & 4,591 (49.6\%) \\
    Pinterest & 1,496 (16.2\%) \\
    Shutterstock & 966 (10.4\%) \\
    Instagram & 933 (10.1\%) \\
    Archdaily & 785 (8.5\%) \\
    Designboom & 477 (5.2\%) \\
    \bottomrule
\end{tabular}}
\end{minipage}
\hspace{0.4\textwidth}  
\begin{minipage}{0.45\textwidth}  
\centering
\includegraphics[width=0.99\linewidth]{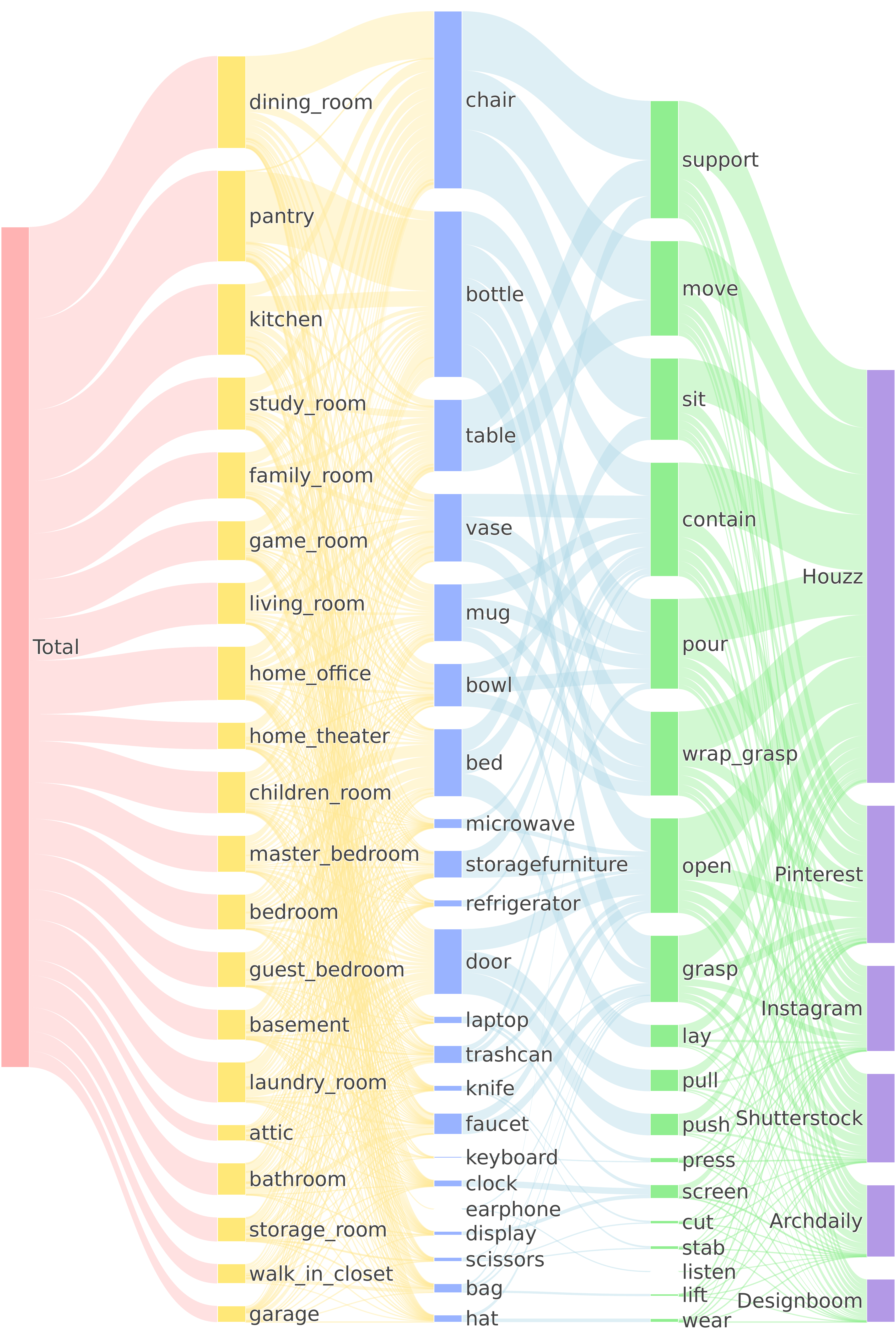}  
\end{minipage}
\caption{\textbf{Dataset overview.} \textbf{(a)} Comprehensive statistics of our dataset, including basic counts, averages, and distribution information. \textbf{(b)} The data collection and processing workflow of our dataset.}
\label{fig:dataset_overview}
\end{figure*}

\subsection{Vision-Language Understanding}
The vision-language understanding process in 3D-TAFS integrates visual and textual inputs through a sophisticated parallel processing architecture. For the visual input, the framework begins with a linear projection of patches from the target image, followed by multi-head attention and add \& norm operations. This is further refined through an MLP layer and another add \& norm step, ultimately producing a visual token. Concurrently, the textual input, which includes an action (e.g., open, pull) and an object (e.g., bottle, door), undergoes its own processing stream. This involves multi-head attention, add \& norm, feed forward, and another add \& norm operation, resulting in a text embedding. Both the visual token and text embedding are then fed into a language model, which performs the crucial task of integrating the visual and linguistic information. This cross-modal attention mechanism allows our framework to establish meaningful connections between the visual elements and the textual instructions. The output of this process is a comprehensive understanding of the scene in relation to the given command, enabling our framework to identify, localize, and classify objects within the image context. This integrated approach forms the foundation for our framework's advanced object understanding and interaction capabilities. However, the specific steps of object localization and classification are not explicitly shown in the diagram.

\subsection{Decision Module and Point Cloud Retrieval}
Based on the output from the vision-language understanding stage, the framework decides whether to proceed to the next step. If the decision is positive, our framework retrieves the corresponding point cloud data from a database, providing 3D spatial information for the identified objects.
\subsection{Language-Guided 3D Affordance Segmentation}
The framework performs language-guided affordance segmentation through a sophisticated multi-step process. The visual input is processed through a series of T-Net layers alternating with add \& norm operations, followed by an MLP layer. Simultaneously, the text input is transformed into embeddings. These parallel streams then converge in a multi-head attention mechanism, allowing the framework to correlate visual features with textual instructions. The resulting fused multimodal representation undergoes further refinement through another MLP layer and a final add \& norm operation. This intricate architecture enables the framework to integrate visual and linguistic information effectively, producing a comprehensive understanding of the scene and instructions that can guide precise, context-aware robotic actions in 3D space.

\setlength{\tabcolsep}{4pt}
\begin{table*}[t]
\fontsize{8pt}{10pt}\selectfont
\centering
\begin{tabular}{lcccc}
\hline
\textbf{Method} & \textbf{mIoU$\uparrow$} & \textbf{AUC$\uparrow$} & \textbf{SIM$\uparrow$} & \textbf{MAE$\downarrow$} \\
\hline
GLIP \cite{Li_2022_CVPR}+ReferTrans  \cite{li2021referring} & 11.1 & 75.0 & 0.412 & 0.136 \\
GDINO \cite{liu2023grounding}+ReferTrans  \cite{li2021referring} & 12.5 & 77.4 & 0.422 & 0.133 \\
NExT-Chat \cite{zhang2024next}+ReferTrans  \cite{li2021referring} & 13.4 & 78.4 & 0.455 & 0.129 \\
GLIP \cite{Li_2022_CVPR}+ReLa \cite{liu2023gres} & 14.0 & 72.5 & 0.502 & 0.124 \\
GDINO \cite{liu2023grounding}+ReLa \cite{liu2023gres} & 14.7 & 75.4 & 0.515 & 0.122 \\
NExT-Chat \cite{zhang2024next}+ReLa \cite{liu2023gres} & 15.5 & 76.6 & 0.524 & 0.119 \\
GLIP \cite{Li_2022_CVPR}+IAGNet \cite{yang2023grounding} & 15.9 & 77.3 & 0.531 & 0.117 \\
GDINO \cite{liu2023grounding}+IAGNet \cite{yang2023grounding} & 16.4 & 79.2 & 0.536 & 0.115 \\
NExT-Chat \cite{zhang2024next}+IAGNet \cite{yang2023grounding} & 17.2 & 80.2 & 0.542 & 0.114 \\
GLIP \cite{Li_2022_CVPR}+PointRefer \cite{Li_2024_CVPR} & 17.4 & 81.6 & 0.547 & 0.113 \\
GDINO  \cite{liu2023grounding}+PointRefer \cite{Li_2024_CVPR} & 18.1 & 81.8 & 0.573 & 0.106 \\
\hline
\textbf{3D-TAFS(NExT-Chat \cite{zhang2024next}+PointRefer \cite{Li_2024_CVPR})} & \textbf{19.1} & \textbf{82.9} & \textbf{0.601} & \textbf{0.099} \\
\hline
\end{tabular}
\caption{\textbf{Performance comparison with state-of-the-art methods.} Our proposed 3D-TAFS achieves the best performance across all metrics: mean Intersection over Union (mIoU) for 3D segmentation accuracy, Area Under the Curve (AUC) for overall performance across different affordance detection thresholds, Similarity (SIM) for predicted and ground truth 3D affordance segmentation comparison, and Mean Absolute Error (MAE) for average magnitude of affordance prediction errors in 3D space.}
\label{tab:performance_comparison}
\end{table*}

Our framework achieves end-to-end mapping from 2D images and text instructions to 3D affordance segmentation through this comprehensive process. The framework's innovation lies in combining powerful vision-language understanding capabilities, intelligent decision-making mechanisms, and precise 3D affordance segmentation techniques, enabling complex language instructions to be directly transformed into operational areas in 3D space.

\section{IndoorAfford-Bench}\label{sec:dataset}
In Figure~\ref{fig:dataset_overview}, we introduce IndoorAfford-Bench, a comprehensive dataset for evaluating interactive language-guided affordance segmentation in everyday environments, building upon the work \cite{Li_2024_CVPR,deng20213d}. Our dataset encompasses 9,248 images across 20 diverse indoor scenes, with rich annotations including 22 object categories and 18 affordance types. These annotations include object labels, affordance segmentation information, and 180 standardized interaction queries generated by GPT4 \cite{achiam2023gpt}.
\subsection{Dataset Structure and Annotation}
Our dataset covers six main categories of indoor spaces. Living Areas include aliving room, family room, and game room, featuring chairs, tables, and entertainment equipment that support social activities. Dining \& Kitchen Areas comprise dining room, kitchen, and pantry spaces, containing essential appliances like microwave and refrigerator. Sleeping Areas contain master bedroom, bedroom, and guest bedroom, focusing on rest and personal storage. Work \& Study Areas include home office, study room, and children's room, equipped with productivity-focused furniture. Storage Areas feature storage room, walk-in closet, basement, and attic, emphasizing organization solutions. Utility Areas comprise bathroom, laundry room, garage, and home theater, each serving specific functional purposes. Each image in the dataset is meticulously annotated with: precise bounding boxes and its label for all relevant objects; detailed affordance segmentation masks for manipulable objects; object-action correspondences listing possible actions for each object; and natural language descriptions of possible tasks and interactions.

\begin{figure*}[t]
  \centering
  \includegraphics[width=0.99\textwidth]{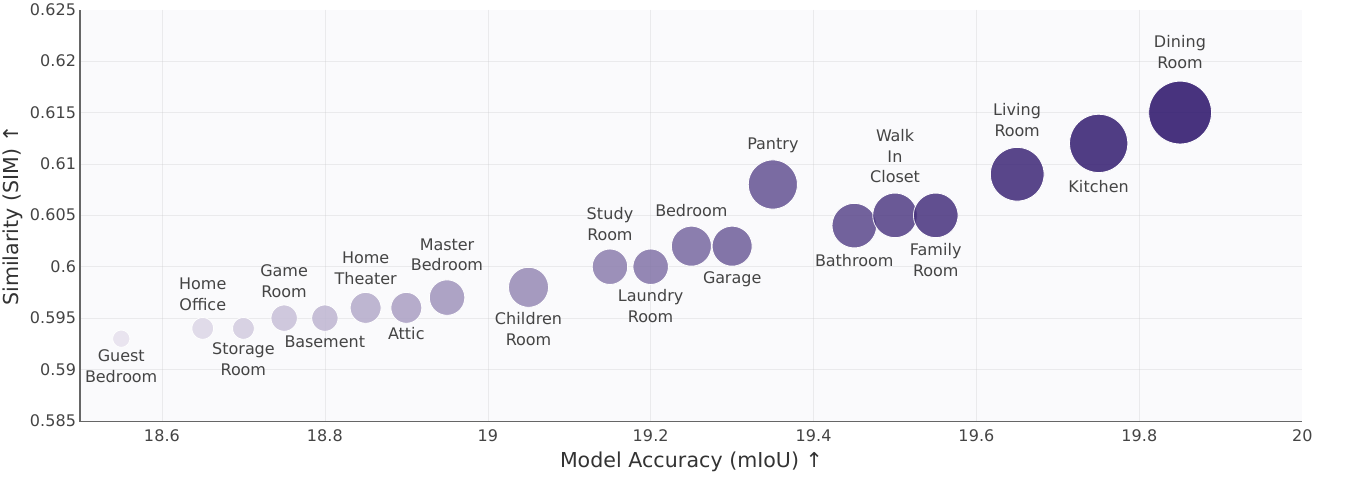}
  \caption{\textbf{ Visualization of room classification performance across different indoor spaces.} The scatter plot displays the relationship between mIoU and SIM, with bubble sizes indicating MAE.  }
  \label{fig:environment_performance_1}
\end{figure*}

\begin{figure}[t]
  \centering
  \includegraphics[width=0.5\textwidth]{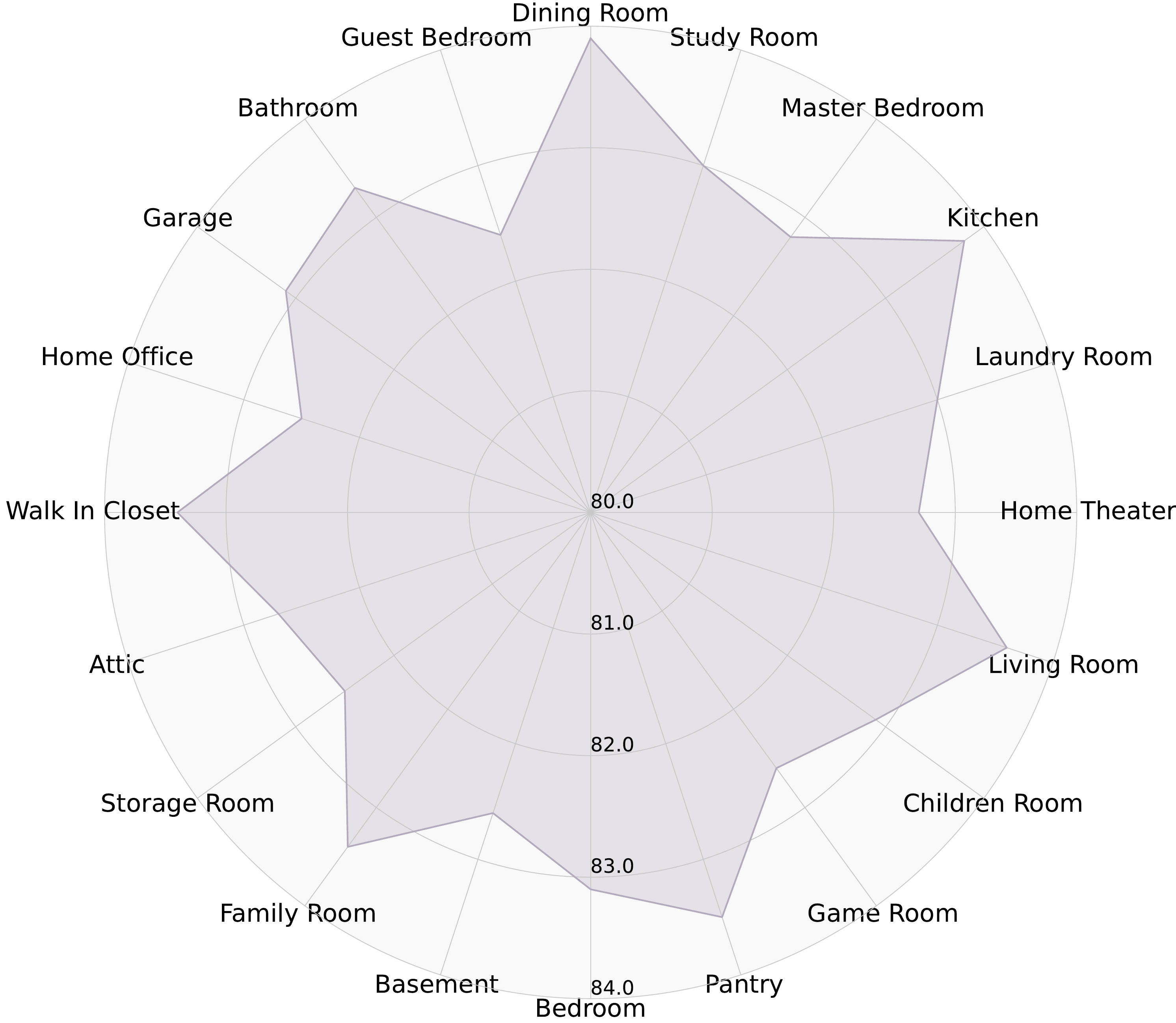}
  \caption{\textbf{ Radar visualization of interactive 3D segmentation performance across different indoor room types with AUC. }  }
  \label{fig:environment_performance_2}
\end{figure}

\subsection{Affordance Annotations}
The dataset provides three main categories of affordance annotations. Basic Interactions include support, move, sit, and contain, covering fundamental object-agent interactions. Manipulation Actions comprise pour, wrap, open, and grasp, representing complex object handling capabilities. Positioning Actions include lay, pull, push, and press, defining spatial manipulation abilities. Each scene averages 167.50 queries and 161.20 queries per image.

\subsection{Data Sources}
IndoorAfford-Bench integrates data from six sources: Houzz (49.6\%), featuring professional interior design photos; Pinterest (16.2\%), providing user-curated content; Shutterstock (10.4\%) and Instagram (10.1\%), offering diverse real-world environments; Archdaily (8.5\%) and Designboom (5.2\%), contributing architectural and design-focused content. This diverse sourcing ensures comprehensive coverage of indoor environments and interaction scenarios.

\section{Experiments} \label{sec:results}

To evaluate the effectiveness of our proposed framework for 3D affordance analysis, we conducted extensive experiments comparing it with state-of-the-art backbones and analyzing its performance across various indoor environments. This section details our experimental setup, quantitative results, qualitative analysis, and discussion of findings.

\subsection{Experimental Setup}

We evaluated 3D-TAFS on a diverse dataset of indoor environments, encompassing ten different room types commonly found in residential settings on one 80G A100 GPU. The dataset includes various objects with various affordances to test the framework's capability in 3D affordance analysis and segmentation. Our experiments were designed to assess both the quantitative performance metrics and qualitative aspects of the framework's understanding and interaction capabilities.

\subsection{Quantitative Results}
\subsubsection{Comparative Analysis}
As shown in Table \ref{tab:performance_comparison}, we compare our framework with other state-of-the-art backbones in 3D affordance analysis and segmentation. 
3D-TAFS outperforms all baseline methods across all metrics, with notable improvements. It achieves 19.1\% in mIoU, surpassing GDINO+PointRefer (18.1\%) by 1\%. The framework reaches an AUC score of 82.9\%, demonstrating superior overall performance. The SIM score of 0.601 shows accurate 3D affordance segmentations, while the lowest MAE of 0.099 highlights precision in affordance localization.

\begin{figure*}[t]
  \centering
  \includegraphics[width=0.63\textwidth]{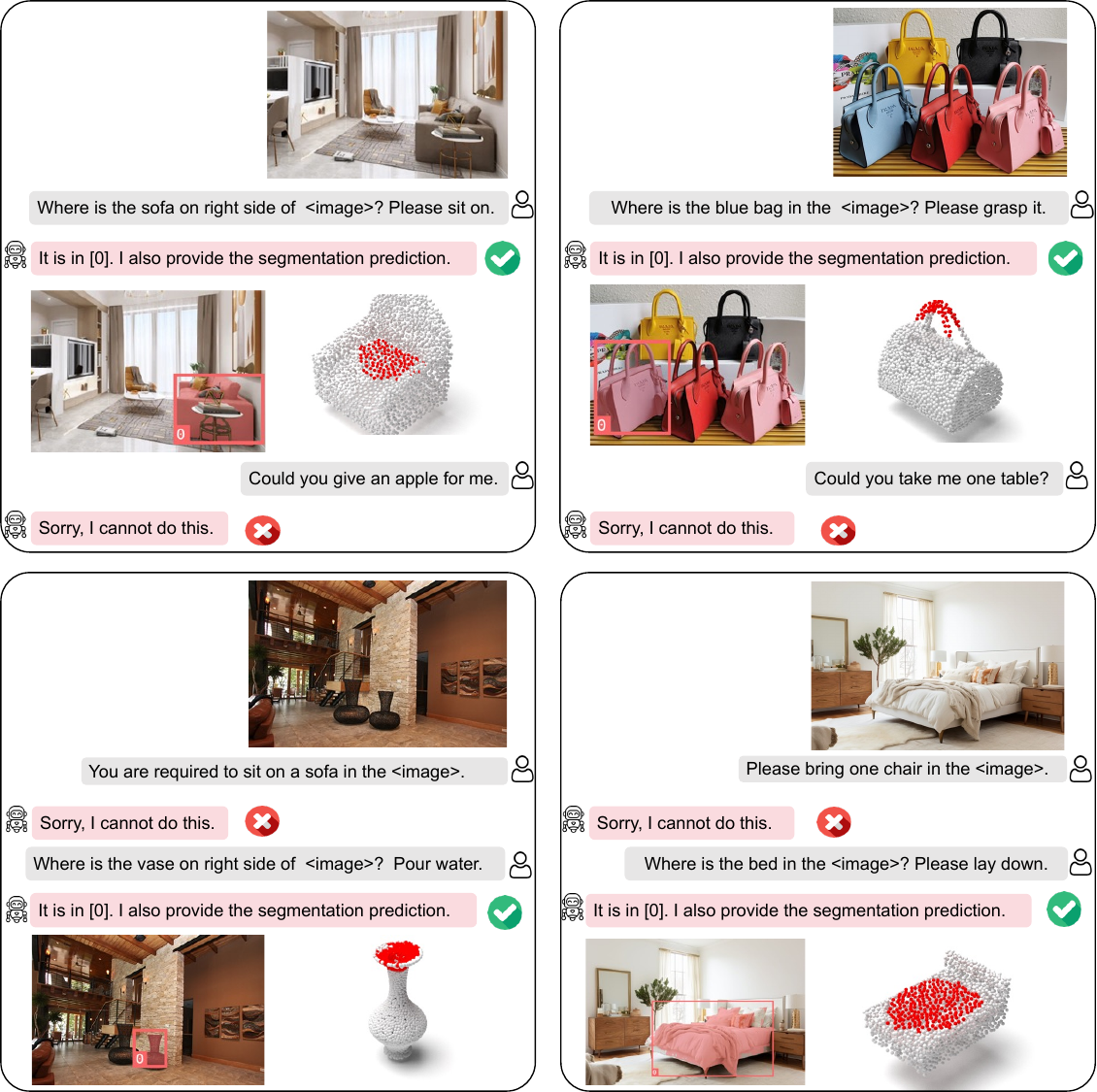}
  \caption{\textbf{Case study of interactive 3D segmentation.} It contains four panels, each featuring an image (such as a living room, handbags, a lobby, and a bedroom) along with questions or instructions about objects in the scenes. The agent successfully answers questions about object locations and provides segmentation predictions when the object exists in the image. }
  \label{fig:show}
\end{figure*}

\subsubsection{Performance Across Different environments}
In Figure~\ref{fig:environment_performance_1} and Figure~\ref{fig:environment_performance_2}, we present 3D-TAFS's performance across twenty indoor environments. Our analysis of indoor scene segmentation reveals distinct performance patterns across different residential spaces. Through both scatter and radar visualizations, we observe that common living areas (dining room, kitchen, living room) consistently achieve higher model accuracy and similarity scores, albeit with larger mean absolute errors, suggesting these spaces are well-recognized but with higher variability in predictions. Conversely, utility spaces (storage room, attic, basement) show lower but more stable performance metrics. This pattern might reflect the inherent complexity of different room types - frequently used spaces contain more diverse features and layouts, leading to better recognition but higher error rates, while utility rooms maintain more standardized characteristics. The radar plot further reinforces these findings, showing balanced performance across most room types with notable variations in segmentation accuracy. This understanding of room-specific performance could guide future improvements in interactive 3D segmentation systems.

\subsection{Qualitative Results}
To complement our quantitative results, we conducted a qualitative analysis of our framework's performance across various scenarios, as illustrated in Figure~\ref{fig:show}. Our framework demonstrates strong performance in diverse environments, accurately identifying and segmenting objects such as a sofa in a living room environment, a blue bag among multiple bags, a vase in an entrance area, and a bed in a bedroom environment. The framework shows a good grasp of object affordances, associating "sit on" with the sofa, recognizing the "grasp" affordance for the blue bag, understanding the "pour water" action for the vase, and correctly interpreting the "lay down" affordance for the bed. Our framework generates accurate 3D point cloud representations of segmented objects, capturing their shape and structure, which is crucial for potential applications in robotics and augmented reality. The qualitative results also reveal our framework's interactive function. It correctly identifies its inability to perform physical actions (e.g., giving an apple or taking a table). It demonstrates an understanding of its role as an analysis and segmentation framework, not a physical actor.

\section{Conclusion} \label{sec:conclusion}

This paper makes two significant contributions to advance 3D affordance segmentation for human-robot interaction. First, we present 3D-TAFS, a novel training-free multimodal framework that effectively bridges linguistic instructions with physical robotic actions through seamless integration of 2D and 3D visual understanding with language comprehension. Second, we introduce IndoorAfford-Bench, a large-scale benchmark containing 9,248 images across 20 diverse indoor scenes, establishing a comprehensive evaluation framework for interactive affordance segmentation tasks. Through extensive experiments on IndoorAfford-Bench, we demonstrate 3D-TAFS's strong performance across various metrics and its capability to handle diverse interaction scenarios. These contributions lay a solid foundation for developing more intuitive and efficient robotic systems capable of understanding and executing complex tasks in real-world indoor environments.

\section*{Limitations}
This study has several limitations that should be acknowledged. As 2D to 3D affordance segmentation is an emerging research direction, we were unable to conduct comparative analyses with existing methods due to the current lack of established baselines in this specific domain. Additionally, while our approach primarily focuses on indoor environments, which encompass a significant portion of robotic applications, it may not fully generalize to other important settings, such as industrial environments where robots are also frequently deployed. Further validation would be needed to assess the model's performance in these different contexts. Although our framework shows promising results in simulation, we have not yet implemented and validated it on physical robotic systems. Real-world deployment would likely introduce additional challenges, such as sensor noise, lighting variations, and real-time processing requirements, that need to be addressed in future work. These limitations present opportunities for future research, including establishing benchmark comparisons, extending the approach to diverse environments, and conducting real-world robotic experiments.

\section*{Ethics Statement}
Our research on 3D affordance understanding was conducted with careful consideration of ethical implications. The dataset was collected in controlled indoor environments with proper consent, ensuring no personal or sensitive information was captured. We acknowledge the potential dual-use nature of affordance detection technology and explicitly prohibit its application in surveillance or harmful scenarios. Our framework is designed for assistive robotics in everyday indoor tasks, aiming to enhance accessibility and independent living while minimizing computational resources. We commit to maintaining transparency about our system's capabilities and limitations and encourage the research community to build upon this work in ways that benefit society.
\bibliography{custom}

\appendix

\end{document}